\documentclass[11pt]{article}
\usepackage{coling2020}
\usepackage{times}
\usepackage{url}
\usepackage{amsmath,amssymb}
\usepackage{latexsym}
\usepackage[size=small]{todonotes}
\usepackage{booktabs}
\usepackage{enumitem}

\newcommand{\ft}[0] {mFastText}
\usepackage{subcaption}
\usepackage{graphicx}
\usepackage{multirow}
\colingfinalcopy % Uncomment this line for the final submission

\title{Probing Multilingual BERT for Genetic and Typological Signals}

\author{Taraka Rama \\
  University of North Texas \\
  {\tt {\small taraka.kasi@gmail.com}} \\ \And
  Lisa Beinborn\\
  VU Amsterdam \\
  {\tt {\small l.m.beinborn@vu.nl}} \\
  \And
  Steffen Eger\\
  TU Darmstadt\\
  {\tt {\small eger@aiphes.tu-darmstadt.de}} \\}

\date{}

\begin{document}
\maketitle
\begin{abstract}
We probe the layers in multilingual BERT (mBERT) for phylogenetic and geographic language signals across 100 languages and compute language distances based on the mBERT representations.  We 1) employ the language distances to infer and evaluate language trees, finding that they are close to the reference family tree in terms of quartet tree distance, 2) perform distance matrix regression analysis, finding that the language distances can be best explained by phylogenetic and worst by structural factors and 3) present a novel measure for measuring diachronic meaning stability (based on cross-lingual representation variability) which correlates significantly with published ranked lists based on linguistic approaches. 
Our results contribute to the nascent field of typological interpretability of cross-lingual text representations.

\end{abstract}

\section{Introduction}

\blfootnote{
    \hspace{-0.65cm}  % space normally used by the marker
    This work is licensed under a Creative Commons 
    Attribution 4.0 International License.
    License details:
    \url{http://creativecommons.org/licenses/by/4.0/}.
}

Cross-lingual text representations have become extremely popular in NLP, since they promise universal text processing in multiple human languages with labeled training data only in a single one. They go back at least to the work of \newcite{klementiev-etal-2012-inducing}, and have seen an exploding number of contributions in recent years. Recent cross-lingual models provide representations for about 100 languages and vary in their training objectives. In offline learning, cross-lingual representations are obtained by projecting independently trained monolingual representations into a shared representational space using bilingual lexical resources \cite{faruqui-dyer-2014-improving,artetxe-etal-2017-learning}. In joint learning \cite{Wang*2020Cross-lingual}, the cross-lingual representations are learned directly, for example as a byproduct of large-scale machine translation \cite{Artetxe:2018_massively}.

As parallel data is scarce for less frequent language pairs, the multilingual BERT model (mBERT) simply trains the BERT architecture \cite{DevlinBERT} on multilingual input from Wikipedia. The cross-lingual signal is thus only learned implicitly because mBERT uses the same representational space independent of the input language. This naive approach yields surprisingly high scores for cross-lingual downstream tasks, but the transfer does not work equally well for all languages.  \newcite{pires-etal-2019-multilingual} show that the performance differences between languages are gradual and that the representational similarity between languages seem to correlate with typological features. These relationships between languages remain opaque in cross-lingual representations and pose a challenge for the evaluation of their adequacy. Evaluations in down-stream tasks are an unreliable approximation because they can often be solved without accounting for deep linguistic knowledge or for interdependencies between subgroups of languages \cite{Liang2020XGLUEAN}. 

While more language-agnostic representations can be beneficial to improve the average performance in task-oriented settings and to smooth the performance differences between high- and low-resource languages  \cite{libovicky2019language,zhao2020inducing}, linguists are more interested in the representational differences between languages. 
The field of computational historical linguistics, for example, examines subtle semantic and syntactic cues to infer phylogenetic relations between languages  \cite{rama2013bchap,jager2014phylogenetic}. Important aspects are the diachronic stability of word meaning  \cite{pagel2007frequency,holman2008explorations} and the analysis of structural properties for inferring deep language relationships \cite{greenhill2010shape,wichmann2007use}. 

Traditionally, these phenomena have been approximated using hand-selected word lists and typological databases. Common ancestors for languages are typically inferred based on cases of shared word meaning and surface form overlap and it can be assumed that these core properties are also captured in large-scale cross-lingual representations to a certain extent.
For example, \newcite{beinborn2019semantic} find that phylogenetic relations between languages can be reconstructed from cross-lingual representations if the training objective optimizes monolingual semantic constraints for each language separately as in the multilingual MUSE model \cite{Conneau:2017}. MUSE is restricted to only 29 frequent languages, however.  
While mBERT is a powerful cross-lingual model covering an order of magnitude more languages (104), a better understanding of the type of signal captured in its representations is needed to assess its applicability as a testbed for cross-lingual or historical linguistic hypotheses.
Our analysis quantifies the representational similarity across languages in mBERT and disentangles it along genetic, geographic, and structural factors.

In general, the urge to improve the interpretability of internal neural representations has become a major research field in recent years. Whereas dense representations of images can be projected back to pixels to facilitate visual inspection, interpreting the linguistic information captured in dense representation of languages is more complex \cite{Alishahi2019Analyzing,conneau-kiela-2018-senteval}. 
Diagnostic classifiers \cite{hupkes2018visualisation}, representational stability analysis \cite{abnar2019blackbox} and indirect visualization techniques \cite{belinkov2019} are only a few examples for newly developed probing techniques. They are used to examine whether the representations capture part-of-speech information \cite{zhang2018language}, syntactic agreement \cite{giulianelli-etal-2018-hood}, speech features \cite{chrupala-etal-2017-representations}, and cognitive cues \cite{wehbe2014aligning}. However, the majority of these interpretability studies focus solely on English. \newcite{krasnowska-kieras-wroblewska-2019-empirical} perform a contrastive analysis of the syntactic interpretability of English and Polish %(sentence) 
representations and \newcite{Eger:2020} probe representations in three lower-resource languages. 
Cross-lingual interpretability research for multiple languages focuses on the ability to transfer representational knowledge across languages for zero-shot semantics \cite{pires-etal-2019-multilingual} and for syntactic phenomena \cite{dhar-bisazza-2018-syntactic}. In this work, we contribute to the nascent field of typological and comparative linguistic interpretability of language representations at scale \cite{kudugunta-etal-2019-investigating} and analyze representations for more than 100 languages.

\textbf{Our contributions}: 
We probe the representations of one of the current most popular cross-lingual models (mBERT) 
and find that mBERT lacks information to perform well on cross-lingual semantic retrieval, but can indeed be used to accurately infer a phylogenetic language tree for 100 languages. Our results indicate that the quality of the induced tree depends on the inference algorithm and might also be the effect of several conflated signals. In order to better disentangle phylogenetic, geographic, and structural factors, we go beyond simple tree comparison and probe language distances inferred from cross-lingual representations by means of multiple regression. We find phylogenetic similarity to be the strongest and structural similarity to be the weakest signal in our experiments. The phylogenetic signal is present across all layers of mBERT.
Our analysis not only contributes to a better interpretation and understanding of mBERT, but may also help explain its cross-lingual behavior in downstream tasks \cite{pires-etal-2019-multilingual}.\footnote{Our code and data are available from \url{https://github.com/PhyloStar/mBertTypology}.}
\section{Related work}\label{sec:related}
Representational distance between two languages refers to the (averaged) differences between model representations for selected concepts in the two languages. Interpretability analyses attempt to disentangle the typological factors that influence the representational distance. In this work, we distinguish between phylogenetic, geographic and structural factors. Two languages are considered to be \emph{phylogenetically} close if they descend from a common ancestor language. \emph{Geographically} close languages are languages which are primarily spoken in regions with a small physical distance on Earth. \emph{Structural} similarity between languages refers to shared syntactic and morphological features. For many languages, the three categories overlap, but they are not necessarily linked. For example, Spanish and Basque are geographically close, but structurally and phylogenetically quite distant. 

Previous approaches differ in the type of cross-lingual representations, the number of languages, and the methodology for determining representational distance and for interpreting the typological signal. 
\newcite{kudugunta-etal-2019-investigating} obtain representations for 102 language pairs (English $\leftrightarrow$ language X) using neural machine translation and then visualize the representations using dimensionality reduction. They explore the visualization qualitatively and find clusters which resemble language families. However, when zooming into the clusters, it becomes evident that a mixture of genetic and geographic factors contributes to the representational distance. For instance, Dravidian and Indo-Aryan languages overlap completely. 

 \newcite{eger-etal-2016-language} induce bilingual vector spaces for 21 Europarl languages and quantify representational distance between languages by averaging over the pairwise similarity between word representations. They find that the differences can be better explained by geographic than by phylogenetic factors. Conversely, \newcite{rabinovich2017found} analyze English translations of sentences in 17 Europarl languages and find that syntactic traces of the native language of the translator can best be explained by language genetics. \newcite{Bjerva2019WhatDL} use the same dataset and train language representations on the linguistic structure of the sentences. They find that the representational distance between languages can be better explained by structural similarity (obtained from dependency trees) than by language genetics.  
Pretrained cross-lingual models are optimized for tasks such as bilingual lexicon induction and machine translation. Even if linguistic information is not explicitly provided during training, recent interpretability research indicates that phylogenetic properties are encoded in the resulting representations. \newcite{beinborn2019semantic} obtain representations for  Swadesh word lists  from the MUSE model \cite{Conneau:2017} which jointly optimizes monolingual and crosslingual semantic constraints. They find that hiearchical clustering over the representational distance between languages yields phylogenetically plausible language trees. Interestingly, they cannot trace the phylogenetic signal in representations from the sentence-based LASER model \cite{Artetxe:2018_massively} which is trained to learn language-neutral representations for machine translation. \newcite{libovicky2019language} analyze representations from mBERT and find that clustering over averaged representations for the 104 languages yields phylogenetically plausible language groups. They argue that mBERT is not language-neutral and that semantic phenomena are not modeled properly across languages. In our analysis, we further quantify the representational distance and disentangle it along phylogenetic, geographic, and structural factors. 

\newcite{bjerva2018phonology} 
train cross-lingual representations in an unsupervised way for different linguistic levels including phonology, morphology, and syntax. They  
use them to infer missing typological features for more than 800 languages. \newcite{malaviya2017learning} also infer missing features in typological databases from cross-lingual representations. Inferring such missing features can be considered a form of probing. Indeed, in contemporaneous work, \newcite{choenni2020does} predict typological properties from representations of four different recent state-of-the-art cross-lingual encoders %for typological properties 
using probing classifiers. We do not use probing classifiers in our work because the choice of classifier and the size of its training data may affect the probing outcomes \cite{Eger:2020}.

Table \ref{table:summary} summarizes selected interpretability approaches analyzing the typological signal in crosslingual representations. The findings for the dominant signal type vary strongly due to different choices for the representational model, the analysis unit, and the number of languages. Our work differs from previous work mainly in terms of the battery of tests probing for genetic and typological signals and the preciseness in teasing apart the different typological components. 

\begin{table*}[!ht]
    \centering
     \small
    \begin{tabular}{lllrl} 
    \toprule
         \textbf{Approach} &  \textbf{Representational Target} & \textbf{Unit} &  \textbf{\# Languages} & \textbf{Dominant Signal} \\ 
         \midrule
         \newcite{malaviya2017learning} & translation &  sentences & 1,017 & structural\\
	\newcite{rabinovich2017found} & syntax &  sentences & 17 & phylogenetic\\
	 \newcite{Bjerva2019WhatDL} & syntax & sentences & 20 & structural\\  
	 \newcite{eger-etal-2016-language} & semantics & words & 21 & geographical\\
	 \newcite{beinborn2019semantic}& semantics & words & 28 &  geographical\\
	 \newcite{libovicky2019language}& semantics & sentences & 104 &  phylogenetic\\
        \bottomrule
    \end{tabular}
    \caption{Summary of related work on typological interpretability of crosslingual representations.}
        \label{table:summary}
\end{table*}

\section{Methodology}\label{sec:matmet}
In the following, we first briefly describe the two cross-lingual embedding spaces analyzed in this work, mBERT and FastText (\S\ref{sec:spaces}). Then, we detail how we compute distances between languages using representations from these spaces (\S\ref{sec:distances}) and concept lists developed in historical linguistics (\S\ref{sec:conceptlist}). Once we have language distances, we infer trees from distance matrices and compare these trees to gold standard phylogenetic trees  (\S\ref{sec:tree_inference}) to evaluate how strong a historical linguistic tree  signal is contained in our cross-lingual representations (\S\ref{sec:exps}).

\subsection{Cross-lingual Representations}\label{sec:spaces}
We compare two different models: mBERT ia a cross-lingual model trained with a language-neutral contextualized objective and FastText stands for static monolingual word representations that have been aligned into a joint multilingual space. 

\paragraph{mBERT} mBERT is based on the multi-layer bidirectional transformer model BERT \cite{DevlinBERT}. It is trained on the task of masked language modeling and next sentence prediction. The base model consists of 12 representational layers. mBERT is trained on the merged Wikipedias of 104 languages, with a shared word-piece vocabulary. It does not use explicit alignments across languages, thus has no mechanism to enforce that translation equivalent word pairs have similar representations. 
Recently, there has been a vivid debate regarding the quality of representations produced by mBERT.  \newcite{pires-etal-2019-multilingual} claim that it is surprisingly good at zero-shot cross-lingual transfer, and works best for structurally similar languages. \newcite{Cao2020Multilingual} find that mBERT exhibits vector space misalignment across languages and zero-shot cross-lingual transfer is improved after their suggested re-mapping. \newcite{K2020Cross-Lingual} show that lexical overlap plays no big role in cross-lingual transfer for mBERT, but the depth of the network does, with deeper models having better transfer.  \newcite{zhao2020limitations} find that mBERT lacks fine-grained cross-lingual text understanding and can be fooled by adversarial inputs produced by the corrupt input produced by MT systems.

\paragraph{FastText} FastText \cite{bojanowski2016enriching} builds static word representations on the basis of a word's characters. This allows it to induce better representations for infrequent and unknown words. We use  a joint multilingually aligned vector space spanning 44 languages using the RCLS method described in \newcite{joulin2018loss} and refer to it as \ft{}.\footnote{\url{https://fasttext.cc/docs/en/aligned-vectors.html}} 

\subsection{Representational Distance}\label{sec:distances}
Assume we have $M$ languages and $N$ concepts (illustrated in Table \ref{tab:swadesh} in the appendix). 
Assume further that each concept is expressed as a word in each language which is represented by a $d$-dimensional vector.

If all the vectors reside in a cross-lingually shared space, then the representational distance between two languages can be obtained by averaging the pairwise distances between all word vectors in the two languages for the $N$ concepts. That means one computes: 
\begin{align}\label{eq:1}
    \text{dist}(i,j) = \frac{1}{N}\sum_{k=1}^N d(\mathbf{v}_k(i),\mathbf{v}_k(j))
\end{align}
where %,
$\mathbf{v}_{k}(i)$ and $\mathbf{v}_k(j)$ stand for the vectors corresponding to the $k$-th concept for languages $i$ and $j$, respectively (with words $v_k(i)$ and $v_k(j)$). 
In our experiments, we use cosine distance, but $d$ may in principle refer to any suitable distance measure, e.g., Euclidean distance or Spearman correlation.\footnote{Note that we compute the average of the distances, while it is also possible to compute the distance of the average representations \cite{libovicky2019language}.}$^,$\footnote{An alternative to this direct comparison of the word vectors is a `second-order' encoding where the representation  $\hat{\mathbf{v}}_k(i)$ for a word is determined by the distances of its vector $\mathbf{v}_k(i)$ to the vectors for the $N$ concepts  \cite{eger-mehler-2016-linearity,beinborn2019semantic}: 

\begin{align*} %\label{eq:_2}
    \hat{\mathbf{v}}_k(i) = \Bigl(d(\mathbf{v}_k(i),\mathbf{v}_n(i))\Bigr)_{n\in[1,\ldots,N]}
\end{align*}
Then, $\hat{\mathbf{v}}_k(\cdot)\in\mathbb{R}^N$ while $\mathbf{v}_k(\cdot)\in\mathbb{R}^d$. These second-order vectors can be used in Eq.~\eqref{eq:1} to replace the original vectors $\mathbf{v}_k$.
}

When the corresponding words for each concept are not available in all languages, but only in one language (e.g., English), \newcite{beinborn2019semantic} instead set $\mathbf{v}_k(i)$ to be the nearest neighbor of the English word for concept $k$ in language $i$. This has the advantage that one can infer language distances without translation data in target languages. A drawback of this approach is that the relation between nearest neighbors in a vector space may not be that of similarity but of relatedness, e.g., \emph{nose} is related to \emph{mouth}, but it is not a synonym (meaning-equivalent). 

In our experiments below, words for all concepts $k$ are available in all languages, thus we do not need to resort to nearest neighbors of the English words.  

\subsection{Concept Lists}\label{sec:conceptlist}
All our experiments are based on multilingual word lists obtained from linguistic databases. 
NorthEuraLex\footnote{\url{http://northeuralex.org/}} features word lists for 1,016 concepts in 100 languages spoken in Northern Eurasia which have been transcribed by linguists \cite{dellert2020LRE}. The database is known for its high quality, but unfortunately covers only 54 of the 104 languages in mBERT. In order to analyze more languages, we additionally use PanLex\footnote{\url{http://dev.panlex.org/db/panlex_swadesh.zip}} which contains lists for 207 concepts in more than 500 languages \cite{kamholz2014panlex}.  It covers 99 languages in mBERT, but the quality of the word lists is not uniform across languages. PanLex sometimes includes multiple word lists written in different scripts for the same language, e.g. for Greek.  
In such a case, we include all available word lists for the language in our analysis.

\subsection{Evaluating language trees}\label{sec:tree_inference}
Historical differences between languages are commonly represented in phylogenetic trees which group languages by their evolution from common ancestors. We want to examine to which extent these phylogenetic differences can explain the observed representational distance in the cross-lingual model. We calculate all pairwise representational distances between languages as in  Eq.~\eqref{eq:1}. From this distance matrix, we infer a language tree using two inference techniques that are widely popular in computational biology for inferring species trees: 
1) The unweighted pair group method with arithmetic mean (UPGMA) \cite{sokal58} initially assumes that each language forms an individual cluster and then successively joins the two clusters with the smallest average distance. 2) The iterative Neighbor Joining \cite{saitou1987neighbor} algorithm starts with an unstructured star-like tree and iteratively adds nodes to create subtrees. 

\paragraph{Reference tree} Most automatic clustering methods produce binary trees due to computational simplifications whereas phylogenetic trees by linguistic experts are usually $m$-ary. To faciliate the evaluation of the inferred tree, previous work used a binary Levenshtein-based approximation \cite{serva2008indo} as reference tree. This approximation provides an acceptable reference for a small subset of languages, but does not accurately reflect the more fine-grained differences for the Indo-European language family \cite{fortson2004}. As we are evaluating a much larger set of languages here, we use the more reliable reference trees compiled by linguistic experts available in Glottolog \cite{glottolog}.

\paragraph{Tree evaluation} In order to compare the $m$-ary reference tree to the binary inferred tree, we apply a variant of quartet distance known as generalized quartet distance \cite{pompei2011accuracy}. This metric evaluates the quality of the whole tree by comparing subgroups of four languages (quartets) which form so-called butterfly structures. A butterfly quartet refers to a quartet in which the four languages can be structured as two pairs of languages belonging to the same subfamily. For example, the pairs Spanish/Italian and Russian/Ukrainian form a butterfly structure whereas the four languages Hindi-German-Armenian-Latin all belong to different subgroups which are directly connected to the root node of the Indo-European tree.  We evaluate our inferred tree by calculating the number of butterfly quartets which deviate from the reference tree normalized by the number of all butterfly quartets in the reference tree. 

 It is also possible to subject the distance matrix to other forms of clustering or dimensionality reduction techniques such as k-nearest neighbor, PCA, or t-sne \cite{maaten2008visualizing}. However, such flat clustering methods do not induce a tree structure and are not directly comparable to the reference trees of language families available in an online repository such as Glottolog \cite{glottolog}. 
\section{Experiments}\label{sec:exps}
In the following, we detail a series of probing experiments with both mBERT and FastText. To extract representations from mBERT, we feed a \emph{single word} $v_k(i)$ from a language $i$ corresponding to concept $k$ into mBERT and extract the corresponding representations $\mathbf{v}_k^{(r)}(i)$ in all layers $r=0,\ldots,12$. We are fully aware that using mBERT in a context-independent way ignores main benefits of the model. We do so in order to leverage concept lists at large scale for a majority of the 100 languages available in mBERT. Otherwise, we would have to experiment with sentence-aligned data, which is available only for much smaller subsets ($<30$) of our languages. Nevertheless, we believe that a good contextual model should also be equipped with good context-independent token representations.

\subsection{Cross-lingual Semantics}
Monolingual language models are commonly evaluated by their ability to model semantic similarity and their performance on downstream tasks. For multilingual models, a suitable evaluation task for lexical semantics is bilingual lexicon induction (BLI). The goal is to take an input word in the source language and retrieve its translation-equivalent in the target language. In a decontextualized setting, multiple targets can be considered to be a correct translation due to the polysemy of words. As the word lists only account for a single correct solution, we cast bilingual lexicon induction as a ranking task. We rank all target words in the concept list based on their representational distance to the source word in the model and evaluate this ranking using the mean reciprocal rank (MRR) as proposed in \newcite{glavas-etal-2019-properly}. The MRR ranges from 0 to 1; a value of 1 indicates that the target is correctly ranked on 1, a value of $1/n$ indicates that the target is on averaged ranked on $n$. 

For the PanLex list of 207 words, the MRR obtained by a random baseline would be 0.03. 
The result of mBERT for the language pair (\texttt{bos},\texttt{hrv}) is almost perfect with an MRR of 0.98, but for pairs of more distant languages the 
BLI 
quality is considerably lower. Overall, the average performance for mBERT (0.16) is five times better than random guessing, but consistently lower than the performance for \ft{} (0.46 on average).\footnote{Only for 12 language pairs, the MRR is higher for mBERT than for \ft{}}
Overall, this shows that mBERT does not properly capture multilingual semantics, a finding that is echoed in some other recent works 
\cite{Cao2020Multilingual,zhao2020limitations}. The apparent reason lies in its naive training process, which does not exploit cross-lingual signals but merely trains on the concatenation of all languages.  Nonetheless, the model  performs surprisingly well in some downstream cross-lingual tasks \cite{pires-etal-2019-multilingual}. In the following experiments, we examine whether the model instead relies on typological properties of languages.  

\subsection{Phylogenetic signal} We perform tree inference using both Neighbor-Joining (NJ) and UPGMA on the concept lists from PanLex ($99$ languages) and NorthEuraLex ($54$ languages). We infer trees for all the layers of mBERT and evaluate the quality of the inferred trees as described in Section \ref{sec:tree_inference}. Table \ref{table:tree_distances} shows that especially the initial-middle and the final layers of mBERT yield a small distance to the gold standard trees.

Overall, however, the strength of the phylogenetic signal varies with respect to the selected concept list and the tree inference algorithm. Interestingly, the results for the PanLex word list are better although this setup covers more languages. UPGMA yields lower distances to the gold tree for both concept lists.
In comparison, the results for \ft{} are considerably worse when using UPGMA (\textgreater 0.5), but comparable when using NJ (around 0.32). It should be noted though that \ft{} covers only 44 languages. 

\begin{table}[!htb]
\small
    \centering
    {
    \begin{tabular}{rr|rrrrrrrrrrrrr|r}
         \toprule
         Method&Word List& 0 & 1 & 2 & 3 & 4 & 5 & 6 & 7 & 8 & 9 & 10 & 11 & 12 & Avg.\  \\ \midrule
         \multirow{2}{*}{UPGMA}&PanLex & .34 & .30 & \textbf{.17} & .18 & .21 & .26 & .28 & .20 & .23 & .21 & .22 & .23 & .21 & .20 \\
         &NorthEuralex & .43 & .29 & \textbf{.26} & .28 & .30 & .31 & .31 & .35 & .34 & .32 & .37 & .34 & .32 & .31 \\ \midrule
         \multirow{2}{*}{NJ}& PanLex & .38 &.31&.30&.30&.26&.31&\textbf{.25}&.32&.32&.32&.35&.34&.30&.30\\
         &NorthEuralex & .41& .36& .35& .32& .32& \textbf{.31}& .31& .32& .32& .32& .32& .32& .40& .37\\
         \bottomrule
    \end{tabular}
    }
    \caption{Distances between the Glottolog reference tree and the phylogenetic tree inferred from mBERT representations from the 12 different layers and the average of all layers.  Generalized quarted distances range between $0$ and $1$, lower distances are better. }
    \label{table:tree_distances}
\end{table}

\begin{figure}[!ht]
    \centering
    \includegraphics[trim=1cm 0.8cm 1cm 1.5cm, clip=true, width=0.8\textwidth,height=0.6\textwidth]{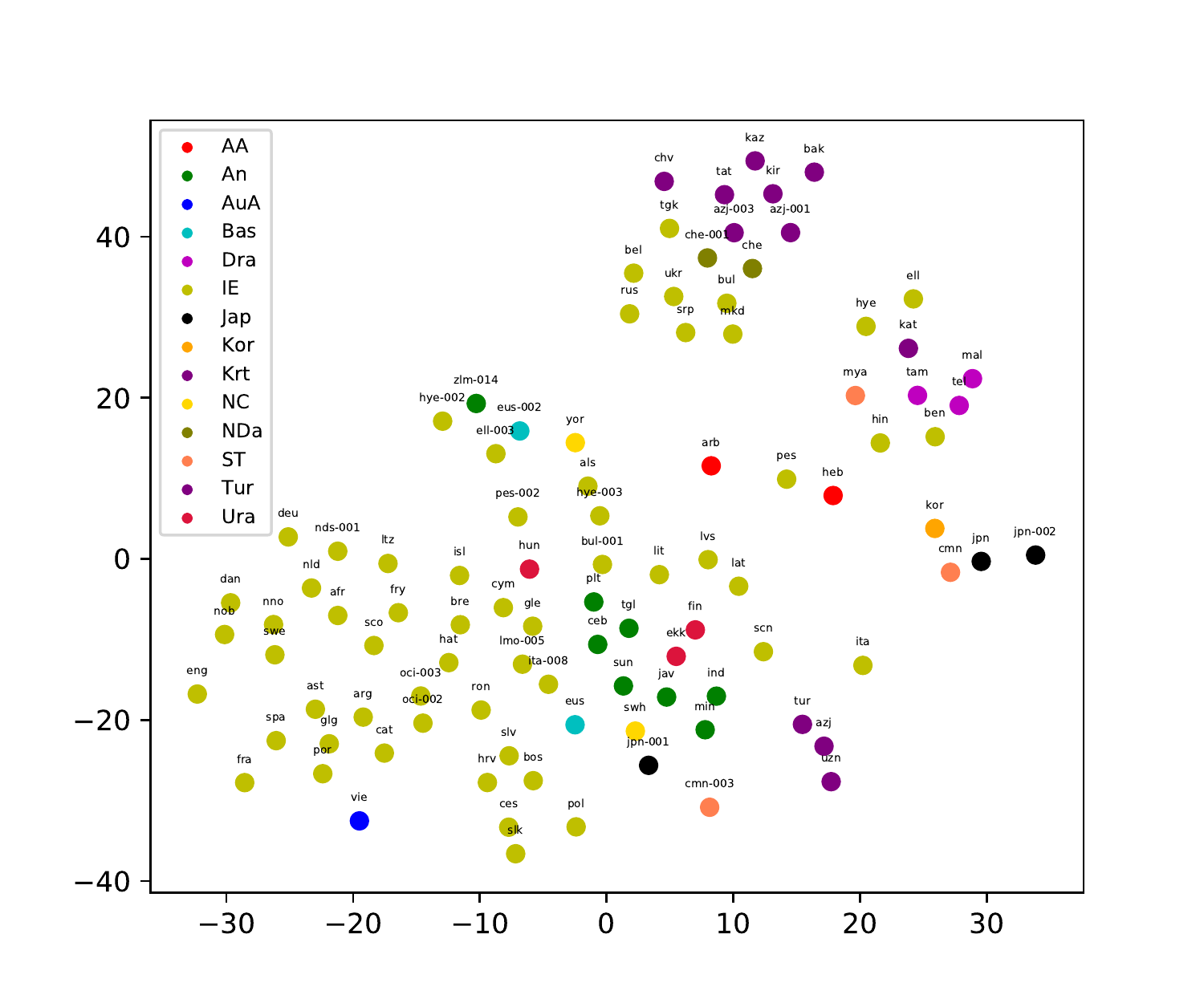}
    \caption{The t-sne plot for the Swadesh list distances from layer 2. The family codes are from the ASJP database \cite{wichmann2020asjp} and are explained in Table \ref{tab:famcodes} in appendix.}
    \label{fig:tsne99}
\end{figure}

\paragraph{Visual exploration}
In order to qualitatively analyze representational distances, we subject the distance matrix from the second layer (which yields the best scores according to UPGMA) to the t-sne algorithm as in previous work  \cite{libovicky2019language,kudugunta-etal-2019-investigating}. The visualization in Figure \ref{fig:tsne99} clearly shows a mix of phylogenetic and other clusters. Instances of phylogenetic clusters include separation of Germanic languages (excluding English which is placed apart) and Romance languages in the lower left. 
In contrast,  the three Dravidian languages (Tamil, Malayalam,Telugu) are placed on the right most part of the plot together with Hindi and Bengali (Indo-European languages),  illustrating more of a geographical similarity.  The Slavic languages show up in two clusters: 1) Western Slavic Languages (Polish, Czech, and Slovak in the lower half) 2) Eastern Slavic languages such as Russian and Ukranian together with Turkic languages such as Azeri and Kazakh written in Cyrllic script. At the same time, the different word lists of Azeri (written in different scripts) are placed together, suggesting that mBERT representations are also script-agnostic. Uralic languages such as Finnish and Estonian are closer to the other Baltic languages which are not clustered together with the other Slavic languages. These clusters cannot be sufficiently explained solely by phylogenetic properties. 

\subsection{Other typological signals}\label{sec:regression}
In order to better disentangle the typological signal, we examine additional categories established by \newcite{littell2017uriel}. We determine the explainable predictors for the representational distances between languages using matrix regression \cite{legendre1994modeling}. We regress $d_{ij}=\text{dist}(i,j)$ computed based on Eq.~\eqref{eq:1} on the following language distances:

\vskip1em
\begin{itemize}[noitemsep]
    \item Phylogenetic distance ($\text{gen}_{ij}$) between two languages computed from Glottolog reference trees as the ratio between the number of non-shared branches divided by the number of branches from root to the tip.
    \item Geographical distance ($\text{geo}_{ij}$) between two points on Earth approximated through great circle distance \cite{great1997admiralty}.
    \item Structural distance ($\text{struc}_{ij}$): Cosine distance computed over averaged syntactic features from WALS\footnote{\url{https://wals.info/}}, SSWL\footnote{\url{http://test.terraling.com/groups/7}}, and mini-grammars parsed from Ethnologue \cite{ethnologue}. 
    \item Phonological distance ($\text{phon}_{ij}$): Cosine distance computed over averaged phonological features available in WALS and Ethnologue.
    \item Phoneme Inventory distance ($\text{inv}_{ij}$): Cosine distance computed between binary feature vectors as given in the PHOIBLE database \cite{phoible} which consist of features such as presence or absence of retroflex sounds in a language.
 \end{itemize}

For detailed description of the distances, see \newcite{littell2017uriel}.\footnote{\url{http://www.cs.cmu.edu/~dmortens/uriel.html}} In our experiments, we use the precomputed distance matrices provided along with the \texttt{lang2vec} Python package\footnote{\url{https://github.com/antonisa/lang2vec}}. We estimate the coefficients as follows: 

\begin{align}\label{eq:regression}
    d_{ij} = c+\alpha\cdot\text{gen}_{ij}+\beta\cdot  \text{geo}_{ij}+\gamma\cdot\text{struc}_{ij} + \eta\cdot\text{phon}_{ij}  + \lambda\cdot\text{inv}_{ij}
\end{align}
Since the entries in the distance matrices are non-independent, we use matrix regression analysis \cite{legendre1994modeling} as implemented in the R package \texttt{ecodist} \cite{goslee2007ecodist} for computing the regression coefficients. The significance of the regression coefficients is also tested using a Mantel test where the matrix columns are permuted $10^5$ times.

\begin{figure}[!ht]
    \centering
    
    \begin{subfigure}{0.48\textwidth}
    \includegraphics[width=\textwidth,height=0.7\textwidth]{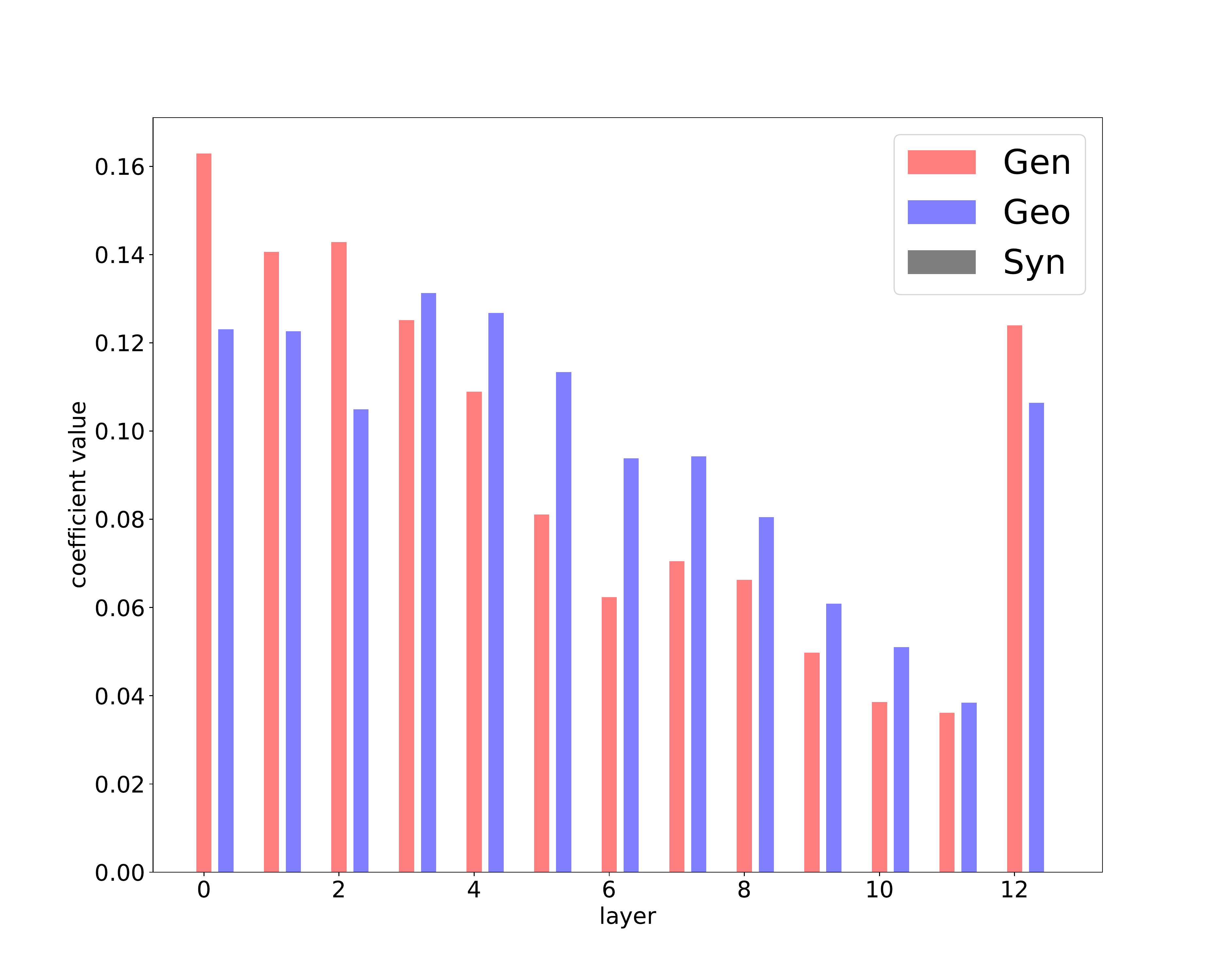}
    \caption{NorthEuralex}
    \label{fig:coefmbertnoreuralex}
    \end{subfigure}
    ~
    \begin{subfigure}{0.48\textwidth}
    \includegraphics[width=\textwidth,height=0.7\textwidth]{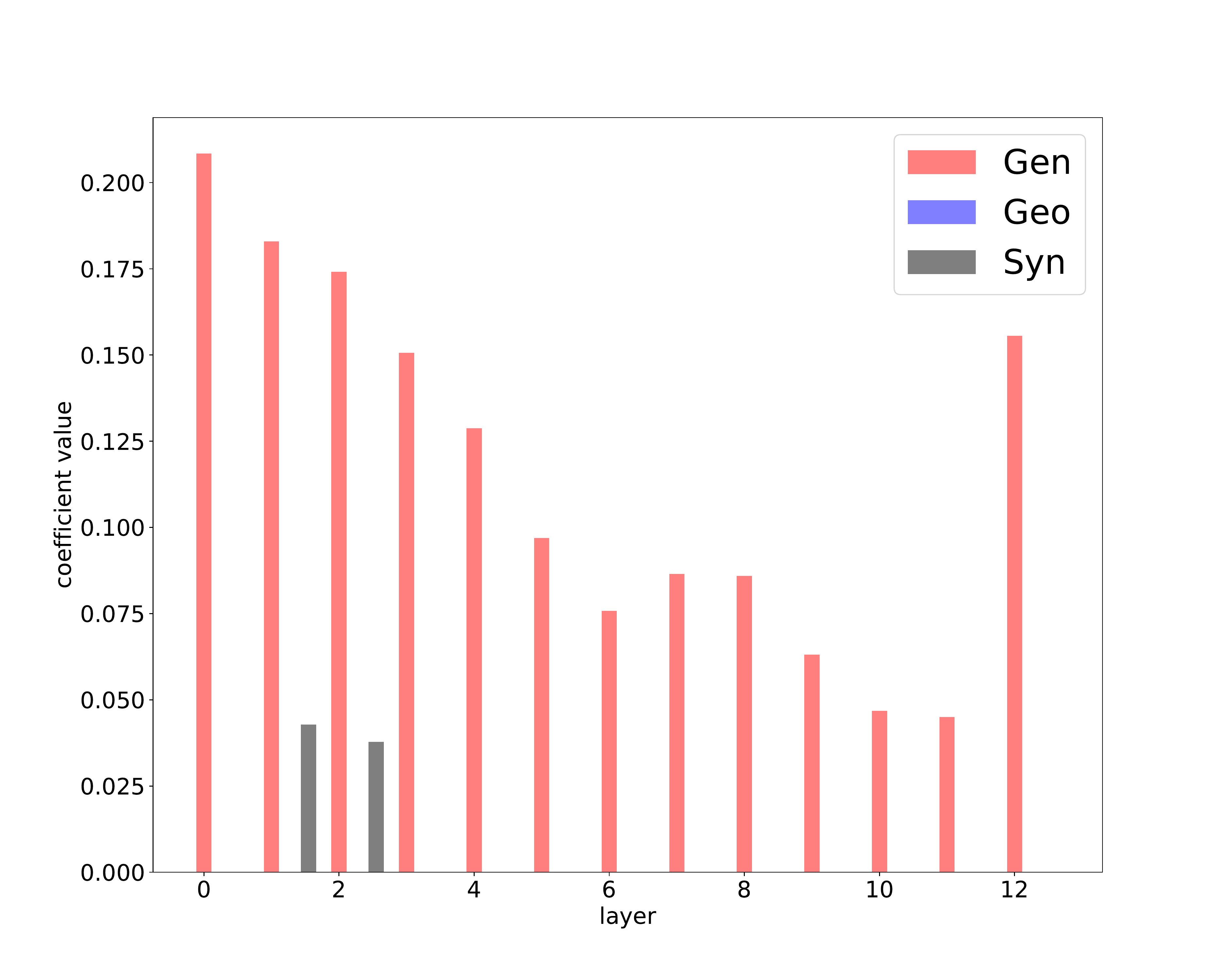}
    \caption{PanLex}
    \label{fig:coefmbertpanlex}
    \end{subfigure}
    
    \caption{The coefficient values ($\alpha, \beta, \gamma$) from Eq.~\eqref{eq:regression} for each mBERT layer.}
    \label{fig:coefmbert}
\end{figure}

The significant regression coefficients ($p < 0.001$) and their sizes are shown in Figure \ref{fig:coefmbert}. The phylogenetic signal in mBERT is stronger than the geographical signal and this is especially true for PanLex where the geographical signal is never significant. The genetic signal is more prominent in the initial layers. It then decreases over layers, but re-emerges in the final layer. As we use isolated concepts in our setup, we did not expect structural features to be a significant predictor at all and are surprised about the PanLex results. 
We further hypothesized that phonological distances might be a weak but significant predictor due to related words (both cognates and borrowings) and shared scripts (also showing up in the t-sne plot in Figure \ref{fig:tsne99}) which is not supported in our experiments. 

Overall, we find that the R\textsuperscript{2} values (see Table \ref{tab:rsquared}) from the regression analyses are significant across all the layers for both the datasets, but they are not very large. 
This suggests that there exist other factors explaining the representational distances that our equation does not account for or that the linear model is not fully appropriate. In the case of the \ft{} model, the R\textsuperscript{2} value is at about $0.24$ but none of the regression coefficients are significant. 

\begin{table}[!htb]
    \centering
    {
    \begin{tabular}{r|rrrrrrrrrrrrr|r}
\toprule
& 0 & 1 & 2 & 3 & 4 & 5 & 6 & 7 & 8 & 9 & 10 & 11 & 12 & Avg.\ \\ \midrule
PanLex & .39 & .39 & \textbf{.41} & .37 & .34 & .28 & .27 & .29 & .29 & .25 & .23 & .25 & .36 & .37\\
NorthEuralex & .39 & .45 & .48 & \textbf{.55} & .54 & .50 & .46 & .43 & .40 & .37 & .36 & .30 & .40 & .48\\
\bottomrule
    \end{tabular}
    }
    \caption{R\textsuperscript{2} values from the regression analyses for each layer.}
    \label{tab:rsquared}
\end{table}

\subsection{Meaning Stability}
We calculate cross-lingual variability %$r$ 
in mBERT by considering the pairwise cosine similarities between representations in languages $i$ and $j$ for all concepts $k$ in PanLex: 
\begin{align}\label{eq:stability}
    c_{ij}(k) = \text{cossim}(\mathbf{v}_k(i),\mathbf{v}_k(j))
\end{align}
and then analyzing how these cosine similarities vary. 
We conjecture that variability of mBERT representations of a concept $k$ across languages is  
indicative of its diachronic stability, where  
diachronic meaning stability measures the resistance of a concept to lexical replacement.

\begin{figure}[!ht]
    \centering
    \begin{subfigure}{0.48\textwidth}
    \includegraphics[width=\textwidth]{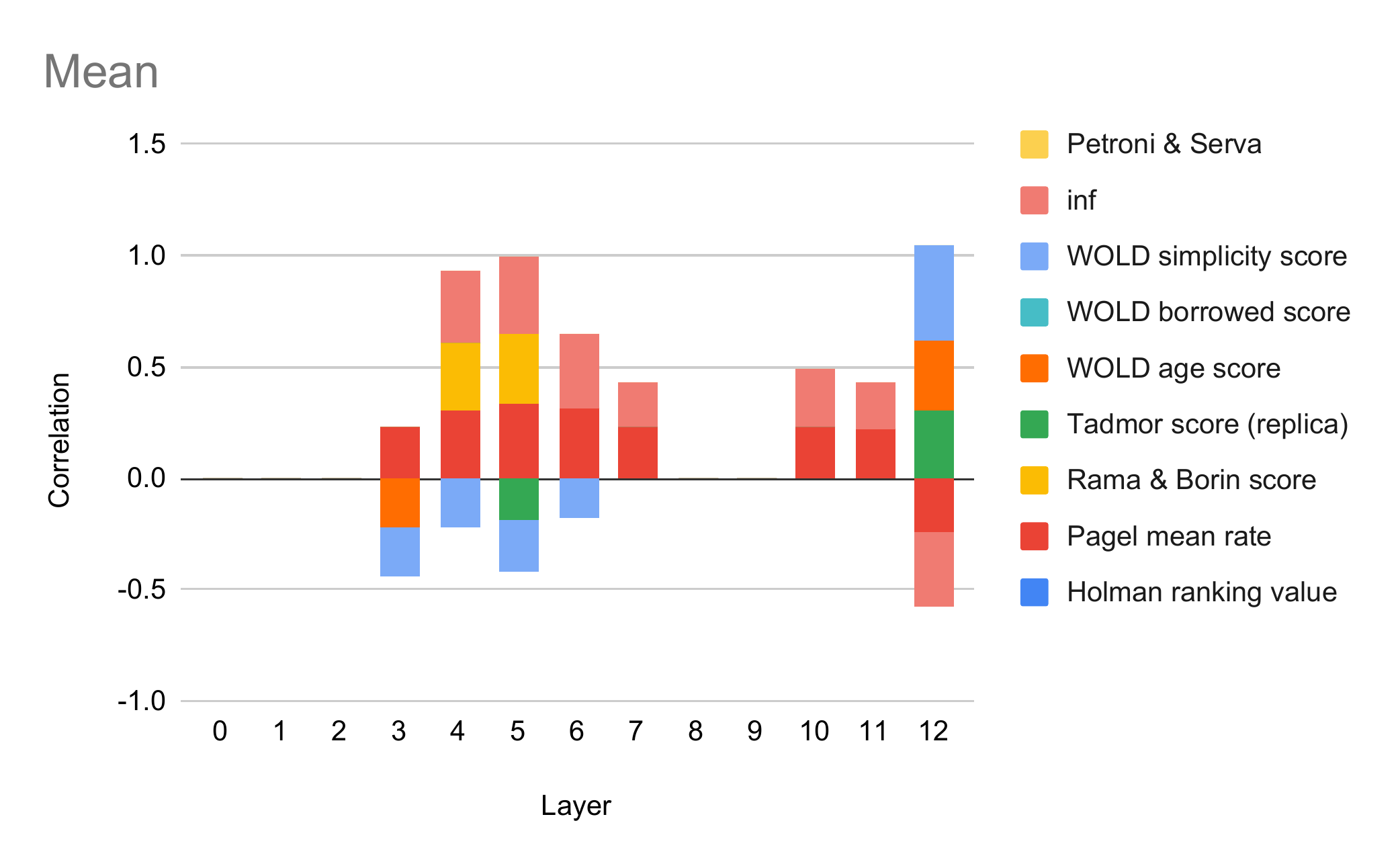}
    \caption{Mean}
    \label{fig:ssmean}
    \end{subfigure}
    ~
    \begin{subfigure}{0.48\textwidth}
    \includegraphics[width=\textwidth]{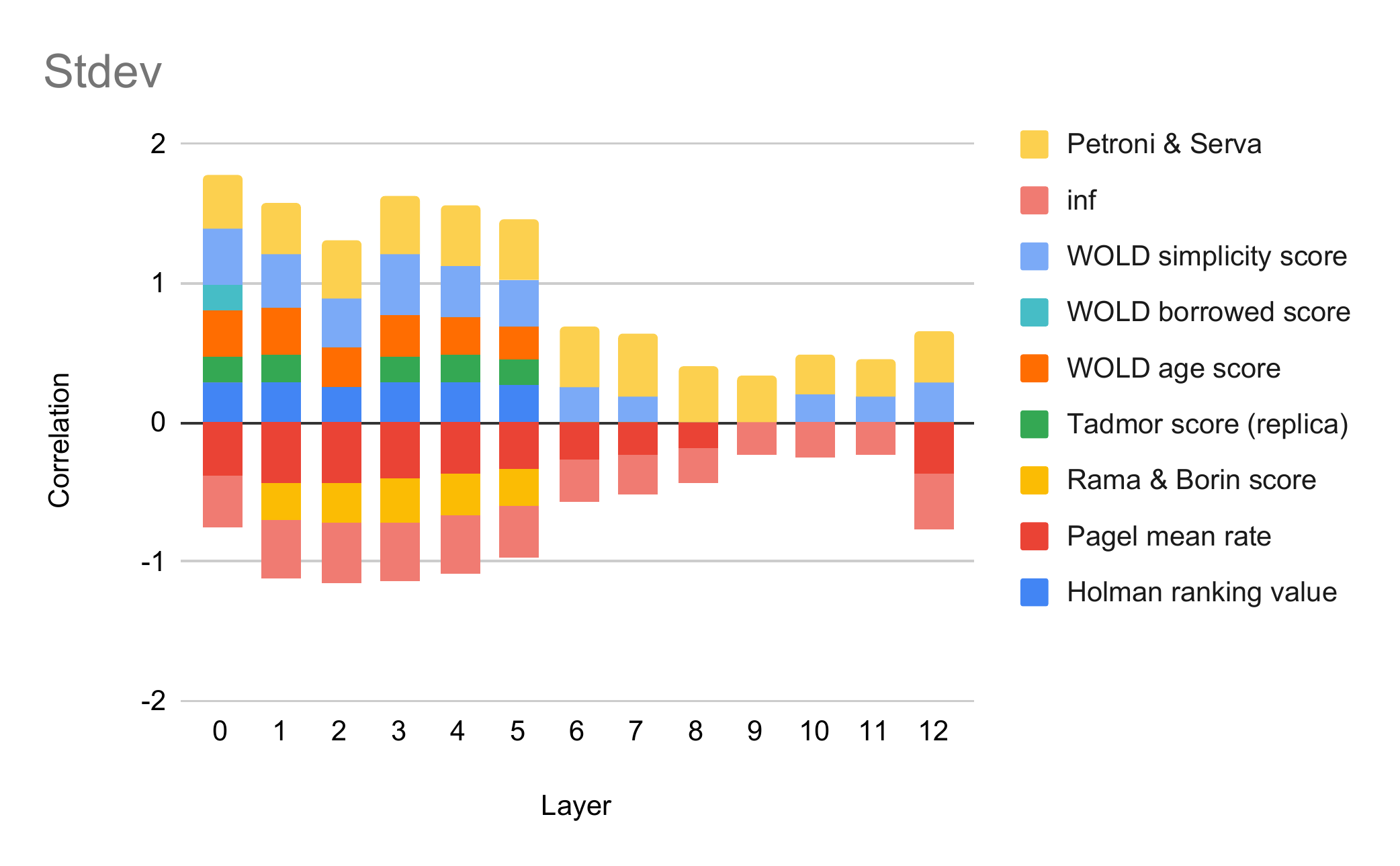}
    \caption{Standard Deviation}
    \label{fig:ssstdev}
    \end{subfigure}
    \caption{
    Correlation between variability of cross-lingual representations from mBERT and diachronic stability lists   
    (significant at $p <.01$).}
    \label{fig:stabrank}
\end{figure}

To capture cross-lingual variability, 
we calculate statistics $s$ on the $c_{ij}(k)$ values. As statistics, we consider the mean and standard deviation values of $c_{ij}(k)$ for each fixed concept $k$. Thus, each concept $k$ receives a ‘variability score’ $s(k)$ given by $s$ applied to the $c_{ij}(k)$ values: $s(k)=s\:(c_{12}(k),c_{13}(k),\ldots,c_{1M}(k),\ldots\:)$, where $M$ is the number of languages involved. 
Note that the standard deviation statistic (intuitively) captures the notion of cross-lingual variability while the mean statistic measures the average degree of similarity of the representations of words for a target concept across languages. 

We finally correlate the statistics with diachronic stability scores for concepts $k$ extracted from the following lists: (i) ranked list of \newcite{holman2008explorations};  (ii) WOLD's (World Loanword Database; \newcite{wold}) meaning scores for age, simplicity, and borrowing; (iii) 100-item Leipzig-Jakarta list \cite{tadmor2009} and its replication (LJ-replica) based on later versions of the WOLD scores \cite{dellert2018new}; (iv) Swadesh list ranked by word replacement rates computed from a phylogenetic analysis \cite{pagel2007frequency}; (v) $n$-gram entropy based stability measure \cite{rama2014n}; (vi) \textit{inf}, a information-theoretic weighted string similarity \cite{dellert2018new}; (vii) and a Levenshtein distance based measure \cite{petroni2011automated}. All the ranked lists are drawn from \newcite{dellert2018new}.

Figure \ref{fig:ssmean} shows the correlation between 
$s(k)$
and diachronic stability when the statistic is the mean. 
It can be seen that Layers 3--7 and 10--11 correlate positively with \newcite{pagel2007frequency} and \emph{inf}. This suggests that the mean statistic measures susceptibility to replacement as opposed to stability since in both the lists, the higher the score, the lower is a concept's stability. Layer 12 correlates with 5 of the 9 lists and shows positive correlation with both WOLD's age and simplicity scores and LJ-replica. 
In contrast, layer 12 correlates negatively with both \newcite{pagel2007frequency} and \emph{inf}, suggesting that the measure is inconsistent in the layer. The mean statistic is also not consistent across the layers and shows inverse correlations with \emph{inf} and \newcite{pagel2007frequency} in layer 12 compared to the other layers.

The correlations from standard deviation statistic is given in Figure \ref{fig:ssstdev}. Here, layers 0--5 show correlations with 8 of the 9 ranked lists. In layers 0--5, the upper half of the figure has positive correlations with word lists ranked by decreasing order of stability (LJ-replica, \newcite{petroni2011automated} and \newcite{holman2008explorations}) whereas the lower half of the figure correlates negatively with the rankings of \newcite{pagel2007frequency}, \emph{inf} and \newcite{rama2014n} where a lower score for a meaning indicates higher stability. Layers 0--7 \& 10--12 correlate positively with WOLD indices such as age and simplicity. There is a negative correlation with \emph{inf} and a positive correlation with the measure of \newcite{petroni2011automated} across all the layers. The standard deviation statistic is consistent across all the layers in terms of correlations against the 9 lists. 

We conclude from this experiment that 
cross-lingual variability of representations in mBERT, as measured by standard deviation,\footnote{Like the mean, other statistics, such as minimum and maximum, did also not exhibit significant correlations. We found significant correlations only for the standard deviation.} indeed correlates with diachronic (cross-temporal) stability as given by proposed historical linguistic indicators. 

\section{Concluding remarks}\label{sec:disc}
We applied a series of tests for probing mBERT for typological signals. While the language trees inferred from mBERT representations are sometimes close to the reference trees, 
they may confound multiple factors. A more-fined grained investigation of t-sne plots followed by matrix regression analyses suggests that representational distances correlate most with phylogenetic and geographical distances between languages. 
Further, the rankings from cross-lingual stability scores correlate significantly with meaning lists for items supposed to be resistant to cross-temporal lexical replacement. 

Our results contribute to the recent discourses on interpretability and introspection of black-box NLP representations \cite{conneau-etal-2018-cram,kudugunta-etal-2019-investigating,jacovi2020faithfully}. In our case, we asked how mBERT perceives of the similarity of two languages and related this to phylogenetic, geographic and structural factors. In future work, we aim to use our inferred similarities to predict transfer behavior in downstream tasks between specific language pairs. 

Finally, we strongly caution against using our conclusions as support for hypotheses relating semantics and language phylogeny (e.g., the Sapir-Whorf hypothesis). Our results for bilingual lexicon induction indicate that mBERT representations are only mildly semantic cross-lingually which corroborates similar findings in related work.
\section*{Acknowledgements}
The last author has been funded by the HMWK (Hessisches Ministerium für Wissenschaft und Kunst) as part of structural location promotion for TU Darmstadt in the context of the Hessian excellence cluster initiative ``Content Analytics for the Social Good'' (CA-SG).
% \input{conclusion}

% include your own bib file like this:
\bibliographystyle{coling}
\bibliography{references}

\begin{thebibliography}{}

\bibitem[\protect\citename{Abnar \bgroup et al.\egroup
  }2019]{abnar2019blackbox}
Samira Abnar, Lisa Beinborn, Rochelle Choenni, and Willem Zuidema.
\newblock 2019.
\newblock Blackbox meets blackbox: Representational similarity and stability
  analysis of neural language models and brains.
\newblock In {\em Proceedings of the ACL-Workshop on Analyzing and Interpreting
  Neural Networks for NLP}, pages 191--203.

\bibitem[\protect\citename{Alishahi \bgroup et al.\egroup
  }2019]{Alishahi2019Analyzing}
Afra Alishahi, Grzegorz Chrupa{\l}a, and Tal Linzen.
\newblock 2019.
\newblock {Analyzing and interpreting neural networks for NLP: A report on the
  first BlackboxNLP workshop}.
\newblock {\em Natural Language Engineering}, 25(4):543--557.

\bibitem[\protect\citename{{Artetxe} and
  {Schwenk}}2018]{Artetxe:2018_massively}
Mikel {Artetxe} and Holger {Schwenk}.
\newblock 2018.
\newblock {Massively Multilingual Sentence Embeddings for Zero-Shot
  Cross-Lingual Transfer and Beyond}.
\newblock {\em arXiv e-prints}, page arXiv:1812.10464.

\bibitem[\protect\citename{Artetxe \bgroup et al.\egroup
  }2017]{artetxe-etal-2017-learning}
Mikel Artetxe, Gorka Labaka, and Eneko Agirre.
\newblock 2017.
\newblock Learning bilingual word embeddings with (almost) no bilingual data.
\newblock In {\em ACL}, pages 451--462.

\bibitem[\protect\citename{Beinborn and Choenni}2019]{beinborn2019semantic}
Lisa Beinborn and Rochelle Choenni.
\newblock 2019.
\newblock Semantic drift in multilingual representations.
\newblock {\em arXiv preprint arXiv:1904.10820}.

\bibitem[\protect\citename{Belinkov and Glass}2019]{belinkov2019}
Yonatan Belinkov and James Glass.
\newblock 2019.
\newblock Analysis methods in neural language processing: A survey.
\newblock {\em Transactions of the Association for Computational Linguistics},
  7:49--72.

\bibitem[\protect\citename{Bjerva and Augenstein}2018]{bjerva2018phonology}
Johannes Bjerva and Isabelle Augenstein.
\newblock 2018.
\newblock From phonology to syntax: Unsupervised linguistic typology at
  different levels with language embeddings.
\newblock In {\em Proceedings of the 2018 Conference of the North American
  Chapter of the Association for Computational Linguistics: Human Language
  Technologies, Volume 1 (Long Papers)}, pages 907--916.

\bibitem[\protect\citename{Bjerva \bgroup et al.\egroup
  }2019]{Bjerva2019WhatDL}
Johannes Bjerva, Robert {\"O}stling, Maria~Han Veiga, J{\"o}rg Tiedemann, and
  Isabelle Augenstein.
\newblock 2019.
\newblock What do language representations really represent?
\newblock {\em Computational Linguistics}, pages 381--389.

\bibitem[\protect\citename{Bojanowski \bgroup et al.\egroup
  }2017]{bojanowski2016enriching}
Piotr Bojanowski, Edouard Grave, Armand Joulin, and Tomas Mikolov.
\newblock 2017.
\newblock Enriching word vectors with subword information.
\newblock {\em TACL}, 5:135--146.

\bibitem[\protect\citename{Cao \bgroup et al.\egroup
  }2020]{Cao2020Multilingual}
Steven Cao, Nikita Kitaev, and Dan Klein.
\newblock 2020.
\newblock Multilingual alignment of contextual word representations.
\newblock In {\em International Conference on Learning Representations}.

\bibitem[\protect\citename{Choenni and Shutova}2020]{choenni2020does}
Rochelle Choenni and Ekaterina Shutova.
\newblock 2020.
\newblock What does it mean to be language-agnostic? probing multilingual
  sentence encoders for typological properties.

\bibitem[\protect\citename{Chrupa{\l}a \bgroup et al.\egroup
  }2017]{chrupala-etal-2017-representations}
Grzegorz Chrupa{\l}a, Lieke Gelderloos, and Afra Alishahi.
\newblock 2017.
\newblock Representations of language in a model of visually grounded speech
  signal.
\newblock In {\em Proceedings of the 55th Annual Meeting of the Association for
  Computational Linguistics (Volume 1: Long Papers)}, pages 613--622,
  Vancouver, Canada, July. Association for Computational Linguistics.

\bibitem[\protect\citename{Conneau and Kiela}2018]{conneau-kiela-2018-senteval}
Alexis Conneau and Douwe Kiela.
\newblock 2018.
\newblock {S}ent{E}val: An evaluation toolkit for universal sentence
  representations.
\newblock In {\em Proceedings of the Eleventh International Conference on
  Language Resources and Evaluation ({LREC} 2018)}, Miyazaki, Japan, May.
  European Language Resources Association (ELRA).

\bibitem[\protect\citename{Conneau \bgroup et al.\egroup }2017]{Conneau:2017}
Alexis Conneau, Douwe Kiela, Holger Schwenk, Lo{\"\i}c Barrault, and Antoine
  Bordes.
\newblock 2017.
\newblock Supervised learning of universal sentence representations from
  natural language inference data.
\newblock In {\em Proceedings of the 2017 Conference on Empirical Methods in
  Natural Language Processing}, pages 670--680, Copenhagen, Denmark, September.
  Association for Computational Linguistics.

\bibitem[\protect\citename{Conneau \bgroup et al.\egroup
  }2018]{conneau-etal-2018-cram}
Alexis Conneau, German Kruszewski, Guillaume Lample, Lo{\"\i}c Barrault, and
  Marco Baroni.
\newblock 2018.
\newblock What you can cram into a single {\$}{\&}!{\#}* vector: Probing
  sentence embeddings for linguistic properties.
\newblock In {\em Proceedings of the 56th Annual Meeting of the Association for
  Computational Linguistics (Volume 1: Long Papers)}, pages 2126--2136,
  Melbourne, Australia, July. Association for Computational Linguistics.

\bibitem[\protect\citename{Dellert and Buch}2018]{dellert2018new}
Johannes Dellert and Armin Buch.
\newblock 2018.
\newblock A new approach to concept basicness and stability as a window to the
  robustness of concept list rankings.
\newblock {\em Language Dynamics and Change}, 8(2):157--181.

\bibitem[\protect\citename{Dellert \bgroup et al.\egroup }2020]{dellert2020LRE}
Johannes Dellert, Thora Daneyko, Alla M{\"{u}}nch, Alina Ladygina, Armin Buch,
  Natalie Clarius, Ilja Grigorjew, Mohamed Balabel, Hizniye~Isabella Boga,
  Zalina Baysarova, Roland M{\"{u}}hlenbernd, Johannes Wahle, and Gerhard
  J{\"{a}}ger.
\newblock 2020.
\newblock {NorthEuraLex: a wide-coverage lexical database of Northern Eurasia}.
\newblock {\em Lang. Resour. Evaluation}, 54(1):273--301.

\bibitem[\protect\citename{Department}1997]{great1997admiralty}
Great Britain.~Navy Department.
\newblock 1997.
\newblock {\em Admiralty Manual of Navigation: BR 45(1)}.
\newblock Number Bd. 1 in BR Series. Stationery Office.

\bibitem[\protect\citename{Devlin \bgroup et al.\egroup }2019]{DevlinBERT}
Jacob Devlin, Ming-Wei Chang, Kenton Lee, and Kristina Toutanova.
\newblock 2019.
\newblock {BERT}: Pre-training of deep bidirectional transformers for language
  understanding.
\newblock In {\em NAACL}, pages 4171--4186, June.

\bibitem[\protect\citename{Dhar and Bisazza}2018]{dhar-bisazza-2018-syntactic}
Prajit Dhar and Arianna Bisazza.
\newblock 2018.
\newblock Does syntactic knowledge in multilingual language models transfer
  across languages?
\newblock In {\em Proceedings of the 2018 {EMNLP} Workshop {B}lackbox{NLP}:
  Analyzing and Interpreting Neural Networks for {NLP}}, pages 374--377,
  Brussels, Belgium, November. Association for Computational Linguistics.

\bibitem[\protect\citename{Eger and Mehler}2016]{eger-mehler-2016-linearity}
Steffen Eger and Alexander Mehler.
\newblock 2016.
\newblock On the linearity of semantic change: Investigating meaning variation
  via dynamic graph models.
\newblock In {\em Proceedings of the 54th Annual Meeting of the Association for
  Computational Linguistics (Volume 2: Short Papers)}, pages 52--58, Berlin,
  Germany, August. Association for Computational Linguistics.

\bibitem[\protect\citename{Eger \bgroup et al.\egroup
  }2016]{eger-etal-2016-language}
Steffen Eger, Armin Hoenen, and Alexander Mehler.
\newblock 2016.
\newblock Language classification from bilingual word embedding graphs.
\newblock In {\em COLING}, pages 3507--3518.

\bibitem[\protect\citename{Eger \bgroup et al.\egroup }2020]{Eger:2020}
Steffen Eger, Johannes Daxenberger, and Iryna Gurevych.
\newblock 2020.
\newblock How to probe sentence embeddings in low-resource languages: On
  structural design choices for probing task evaluation.
\newblock In {\em CONLL}.

\bibitem[\protect\citename{Faruqui and Dyer}2014]{faruqui-dyer-2014-improving}
Manaal Faruqui and Chris Dyer.
\newblock 2014.
\newblock Improving vector space word representations using multilingual
  correlation.
\newblock In {\em Proceedings of the 14th Conference of the {E}uropean Chapter
  of the Association for Computational Linguistics}, pages 462--471,
  Gothenburg, Sweden, April. Association for Computational Linguistics.

\bibitem[\protect\citename{Fortson}2004]{fortson2004}
Benjamin~F. Fortson, IV.
\newblock 2004.
\newblock {\em Indo-European language and culture: an introduction}, volume~19
  of {\em Blackwell Textbooks in Linguistics}.
\newblock Blackwell, Oxford.

\bibitem[\protect\citename{Giulianelli \bgroup et al.\egroup
  }2018]{giulianelli-etal-2018-hood}
Mario Giulianelli, Jack Harding, Florian Mohnert, Dieuwke Hupkes, and Willem
  Zuidema.
\newblock 2018.
\newblock Under the hood: Using diagnostic classifiers to investigate and
  improve how language models track agreement information.
\newblock In {\em Proceedings of the 2018 {EMNLP} Workshop {B}lackbox{NLP}:
  Analyzing and Interpreting Neural Networks for {NLP}}, pages 240--248,
  Brussels, Belgium, November. Association for Computational Linguistics.

\bibitem[\protect\citename{Glava{\v{s}} \bgroup et al.\egroup
  }2019]{glavas-etal-2019-properly}
Goran Glava{\v{s}}, Robert Litschko, Sebastian Ruder, and Ivan Vuli{\'c}.
\newblock 2019.
\newblock How to (properly) evaluate cross-lingual word embeddings: On strong
  baselines, comparative analyses, and some misconceptions.
\newblock In {\em Proceedings of the 57th Annual Meeting of the Association for
  Computational Linguistics}, pages 710--721, Florence, Italy, July.
  Association for Computational Linguistics.

\bibitem[\protect\citename{Goslee \bgroup et al.\egroup
  }2007]{goslee2007ecodist}
Sarah~C Goslee, Dean~L Urban, et~al.
\newblock 2007.
\newblock The ecodist package for dissimilarity-based analysis of ecological
  data.
\newblock {\em Journal of Statistical Software}, 22(7):1--19.

\bibitem[\protect\citename{Greenhill \bgroup et al.\egroup
  }2010]{greenhill2010shape}
Simon~J Greenhill, Quentin~D Atkinson, Andrew Meade, and Russell~D Gray.
\newblock 2010.
\newblock The shape and tempo of language evolution.
\newblock {\em Proceedings of the Royal Society B: Biological Sciences},
  277(1693):2443--2450.

\bibitem[\protect\citename{Hammarström \bgroup et al.\egroup }2020]{glottolog}
Harald Hammarström, Robert Forkel, Martin Haspelmath, and Sebastian Bank.
\newblock 2020.
\newblock Glottolog 4.2.1.
\newblock Max Planck Institute for the Science of Human History.

\bibitem[\protect\citename{Haspelmath and Tadmor}2009]{wold}
Martin Haspelmath and Uri Tadmor, editors.
\newblock 2009.
\newblock {\em WOLD}.
\newblock Max Planck Institute for Evolutionary Anthropology, Leipzig.

\bibitem[\protect\citename{Holman \bgroup et al.\egroup
  }2008]{holman2008explorations}
Eric~W Holman, S{\o}ren Wichmann, Cecil~H Brown, Viveka Velupillai, Andr{\'e}
  M{\"u}ller, and Dik Bakker.
\newblock 2008.
\newblock Explorations in automated language classification.
\newblock {\em Folia Linguistica}, 42(3-4):331--354.

\bibitem[\protect\citename{Hupkes \bgroup et al.\egroup
  }2018]{hupkes2018visualisation}
Dieuwke Hupkes, Sara Veldhoen, and Willem Zuidema.
\newblock 2018.
\newblock Visualisation and 'diagnostic classifiers' reveal how recurrent and
  recursive neural networks process hierarchical structure.
\newblock {\em Journal of Artificial Intelligence Research}, 61:907--926.

\bibitem[\protect\citename{Jacovi and Goldberg}2020]{jacovi2020faithfully}
Alon Jacovi and Yoav Goldberg.
\newblock 2020.
\newblock Towards faithfully interpretable nlp systems: How should we define
  and evaluate faithfulness?

\bibitem[\protect\citename{J{\"a}ger}2014]{jager2014phylogenetic}
Gerhard J{\"a}ger.
\newblock 2014.
\newblock Phylogenetic inference from word lists using weighted alignment with
  empirically determined weights.
\newblock In {\em Language Dynamics and Change}, pages 155--204. Brill.

\bibitem[\protect\citename{Joulin \bgroup et al.\egroup }2018]{joulin2018loss}
Armand Joulin, Piotr Bojanowski, Tomas Mikolov, Herv\'e J\'egou, and Edouard
  Grave.
\newblock 2018.
\newblock Loss in translation: Learning bilingual word mapping with a retrieval
  criterion.
\newblock In {\em Proceedings of the 2018 Conference on Empirical Methods in
  Natural Language Processing}.

\bibitem[\protect\citename{K \bgroup et al.\egroup }2020]{K2020Cross-Lingual}
Karthikeyan K, Zihan Wang, Stephen Mayhew, and Dan Roth.
\newblock 2020.
\newblock Cross-lingual ability of multilingual bert: An empirical study.
\newblock In {\em International Conference on Learning Representations}.

\bibitem[\protect\citename{Kamholz \bgroup et al.\egroup
  }2014]{kamholz2014panlex}
David Kamholz, Jonathan Pool, and Susan~M Colowick.
\newblock 2014.
\newblock Panlex: Building a resource for panlingual lexical translation.
\newblock In {\em LREC}, pages 3145--3150.

\bibitem[\protect\citename{Klementiev \bgroup et al.\egroup
  }2012]{klementiev-etal-2012-inducing}
Alexandre Klementiev, Ivan Titov, and Binod Bhattarai.
\newblock 2012.
\newblock Inducing crosslingual distributed representations of words.
\newblock In {\em Proceedings of {COLING} 2012}, pages 1459--1474, Mumbai,
  India, December. The COLING 2012 Organizing Committee.

\bibitem[\protect\citename{Krasnowska-Kiera{\'s} and
  Wr{\'o}blewska}2019]{krasnowska-kieras-wroblewska-2019-empirical}
Katarzyna Krasnowska-Kiera{\'s} and Alina Wr{\'o}blewska.
\newblock 2019.
\newblock Empirical linguistic study of sentence embeddings.
\newblock In {\em ACL}, pages 5729--5739.

\bibitem[\protect\citename{Kudugunta \bgroup et al.\egroup
  }2019]{kudugunta-etal-2019-investigating}
Sneha Kudugunta, Ankur Bapna, Isaac Caswell, and Orhan Firat.
\newblock 2019.
\newblock Investigating multilingual {NMT} representations at scale.
\newblock In {\em Proceedings of the 2019 Conference on Empirical Methods in
  Natural Language Processing and the 9th International Joint Conference on
  Natural Language Processing (EMNLP-IJCNLP)}, pages 1565--1575, Hong Kong,
  China, November. Association for Computational Linguistics.

\bibitem[\protect\citename{Legendre \bgroup et al.\egroup
  }1994]{legendre1994modeling}
Pierre Legendre, Fran{\c{c}}ois-Joseph Lapointe, and Philippe Casgrain.
\newblock 1994.
\newblock Modeling brain evolution from behavior: a permutational regression
  approach.
\newblock {\em Evolution}, 48(5):1487--1499.

\bibitem[\protect\citename{Lewis}2009]{ethnologue}
M.~Paul Lewis, editor.
\newblock 2009.
\newblock {\em Ethnologue: Languages of the World}.
\newblock SIL International, Dallas, TX, USA, sixteenth edition.

\bibitem[\protect\citename{Liang \bgroup et al.\egroup }2020]{Liang2020XGLUEAN}
Yaobo Liang, Nan Duan, Yeyun Gong, Ning Wu, Fenfei Guo, Weizhen Qi, Ming Gong,
  Linjun Shou, Daxin Jiang, Guihong Cao, Xiaodong Fan, Ruofei Zhang, Rahul
  Agrawal, Edward Cui, Sining Wei, Taroon Bharti, Ying Qiao, Jiun-Hung Chen,
  Winnie Wu, Shuguang Liu, Fan Yang, Daniel Campos, Rangan Majumder, and Ming
  Zhou.
\newblock 2020.
\newblock Xglue: A new benchmark dataset for cross-lingual pre-training,
  understanding and generation.
\newblock {\em arXiv}, abs/2004.01401.

\bibitem[\protect\citename{Libovick{\`y} \bgroup et al.\egroup
  }2019]{libovicky2019language}
Jind{\v{r}}ich Libovick{\`y}, Rudolf Rosa, and Alexander Fraser.
\newblock 2019.
\newblock How language-neutral is multilingual bert?
\newblock {\em arXiv preprint arXiv:1911.03310}.

\bibitem[\protect\citename{Littell \bgroup et al.\egroup
  }2017]{littell2017uriel}
Patrick Littell, David~R Mortensen, Ke~Lin, Katherine Kairis, Carlisle Turner,
  and Lori Levin.
\newblock 2017.
\newblock Uriel and lang2vec: Representing languages as typological,
  geographical, and phylogenetic vectors.
\newblock In {\em Proceedings of the 15th Conference of the European Chapter of
  the Association for Computational Linguistics: Volume 2, Short Papers}, pages
  8--14.

\bibitem[\protect\citename{Maaten and Hinton}2008]{maaten2008visualizing}
Laurens van~der Maaten and Geoffrey Hinton.
\newblock 2008.
\newblock Visualizing data using t-sne.
\newblock {\em Journal of machine learning research}, 9(Nov):2579--2605.

\bibitem[\protect\citename{Malaviya \bgroup et al.\egroup
  }2017]{malaviya2017learning}
Chaitanya Malaviya, Graham Neubig, and Patrick Littell.
\newblock 2017.
\newblock Learning language representations for typology prediction.
\newblock In {\em Proceedings of the 2017 Conference on Empirical Methods in
  Natural Language Processing}, pages 2529--2535.

\bibitem[\protect\citename{Moran and McCloy}2019]{phoible}
Steven Moran and Daniel McCloy, editors.
\newblock 2019.
\newblock {\em PHOIBLE 2.0}.
\newblock Max Planck Institute for the Science of Human History, Jena.

\bibitem[\protect\citename{Pagel \bgroup et al.\egroup
  }2007]{pagel2007frequency}
Mark Pagel, Quentin~D Atkinson, and Andrew Meade.
\newblock 2007.
\newblock Frequency of word-use predicts rates of lexical evolution throughout
  indo-european history.
\newblock {\em Nature}, 449(7163):717--720.

\bibitem[\protect\citename{Petroni and Serva}2011]{petroni2011automated}
Filippo Petroni and Maurizio Serva.
\newblock 2011.
\newblock Automated word stability and language phylogeny.
\newblock {\em Journal of Quantitative Linguistics}, 18(1):53--62.

\bibitem[\protect\citename{Pires \bgroup et al.\egroup
  }2019]{pires-etal-2019-multilingual}
Telmo Pires, Eva Schlinger, and Dan Garrette.
\newblock 2019.
\newblock How multilingual is multilingual {BERT}?
\newblock In {\em Proceedings of the 57th Annual Meeting of the Association for
  Computational Linguistics}, pages 4996--5001, Florence, Italy, July.
  Association for Computational Linguistics.

\bibitem[\protect\citename{Pompei \bgroup et al.\egroup
  }2011]{pompei2011accuracy}
Simone Pompei, Vittorio Loreto, and Francesca Tria.
\newblock 2011.
\newblock On the accuracy of language trees.
\newblock {\em PloS one}, 6(6).

\bibitem[\protect\citename{Rabinovich \bgroup et al.\egroup
  }2017]{rabinovich2017found}
Ella Rabinovich, Noam Ordan, and Shuly Wintner.
\newblock 2017.
\newblock Found in translation: Reconstructing phylogenetic language trees from
  translations.
\newblock In {\em Proceedings of the 55th Annual Meeting of the Association for
  Computational Linguistics (Volume 1: Long Papers)}, pages 530--540.

\bibitem[\protect\citename{Rama and Borin}2014]{rama2014n}
Taraka Rama and Lars Borin.
\newblock 2014.
\newblock N-gram approaches to the historical dynamics of basic vocabulary.
\newblock {\em Journal of Quantitative Linguistics}, 21(1):50--64.

\bibitem[\protect\citename{Rama and Borin}2015]{rama2013bchap}
Taraka Rama and Lars Borin.
\newblock 2015.
\newblock Comparative evaluation of string similarity measures for automatic
  language classification.
\newblock In Ján Mačutek and George~K. Mikros, editors, {\em Sequences in
  Language and Text}, pages 203--231. Walter de Gruyter.

\bibitem[\protect\citename{Saitou and Nei}1987]{saitou1987neighbor}
Naruya Saitou and Masatoshi Nei.
\newblock 1987.
\newblock The neighbor-joining method: a new method for reconstructing
  phylogenetic trees.
\newblock {\em Molecular biology and evolution}, 4(4):406--425.

\bibitem[\protect\citename{Serva and Petroni}2008]{serva2008indo}
Maurizio Serva and Filippo Petroni.
\newblock 2008.
\newblock Indo-european languages tree by levenshtein distance.
\newblock {\em EPL (Europhysics Letters)}, 81(6):68005.

\bibitem[\protect\citename{Sokal and Michener}1958]{sokal58}
R.~R. Sokal and C.~D. Michener.
\newblock 1958.
\newblock A statistical method for evaluating systematic relationships.
\newblock {\em University of Kansas Science Bulletin}, 38:1409--1438.

\bibitem[\protect\citename{Tadmor}2009]{tadmor2009}
Uri Tadmor.
\newblock 2009.
\newblock Loanwords in the world's languages. findings and results.
\newblock In Martin Haspelmath and Uri Tadmor, editors, {\em Loanwords in the
  world's languages. A comparative handbook}, pages 55--75. de Gruyter, Berlin
  and New York.

\bibitem[\protect\citename{Wang \bgroup et al.\egroup
  }2020]{Wang*2020Cross-lingual}
Zirui Wang, Jiateng Xie, Ruochen Xu, Yiming Yang, Graham Neubig, and Jaime~G.
  Carbonell.
\newblock 2020.
\newblock Cross-lingual alignment vs joint training: A comparative study and a
  simple unified framework.
\newblock In {\em International Conference on Learning Representations}.

\bibitem[\protect\citename{Wehbe \bgroup et al.\egroup
  }2014]{wehbe2014aligning}
Leila Wehbe, Ashish Vaswani, Kevin Knight, and Tom Mitchell.
\newblock 2014.
\newblock Aligning context-based statistical models of language with brain
  activity during reading.
\newblock In {\em Proceedings of the 2014 Conference on Empirical Methods in
  Natural Language Processing ({EMNLP})}, pages 233--243, Doha, Qatar, October.
  Association for Computational Linguistics.

\bibitem[\protect\citename{Wichmann and Saunders}2007]{wichmann2007use}
S{\o}ren Wichmann and Arpiar Saunders.
\newblock 2007.
\newblock How to use typological databases in historical linguistic research.
\newblock {\em Diachronica}, 24(2):373--404.

\bibitem[\protect\citename{Wichmann \bgroup et al.\egroup
  }2020]{wichmann2020asjp}
S{\o}ren Wichmann, Eric~W Holman, and Cecil~H Brown.
\newblock 2020.
\newblock {The ASJP database (version 19)}.
\newblock {\em Jena: Max Planck Institute for the Science of Human History}.

\bibitem[\protect\citename{Zhang and Bowman}2018]{zhang2018language}
Kelly Zhang and Samuel Bowman.
\newblock 2018.
\newblock Language modeling teaches you more than translation does: Lessons
  learned through auxiliary syntactic task analysis.
\newblock In {\em Proceedings of the 2018 EMNLP Workshop BlackboxNLP: Analyzing
  and Interpreting Neural Networks for NLP}, pages 359--361.

\bibitem[\protect\citename{Zhao \bgroup et al.\egroup }2020a]{zhao2020inducing}
Wei Zhao, Steffen Eger, Johannes Bjerva, and Isabelle Augenstein.
\newblock 2020a.
\newblock Inducing language-agnostic multilingual representations.

\bibitem[\protect\citename{Zhao \bgroup et al.\egroup
  }2020b]{zhao2020limitations}
Wei Zhao, Goran Glavaš, Maxime Peyrard, Yang Gao, Robert West, and Steffen
  Eger.
\newblock 2020b.
\newblock On the limitations of cross-lingual encoders as exposed by
  reference-free machine translation evaluation.
\newblock In {\em ACL}.

\end{thebibliography}
\newpage

\appendix

\section{Example list}

\begin{table}[!htb]
  \centering
  \begin{tabular}{l|cccc}
       \toprule
       & English & German & French & $\cdots$ \\ \midrule
       I &  I & ich & je & $\cdots$ \\
       YOU & you & du & tu & $\cdots$\\
       HE & he & er & il & $\cdots$\\
       MAN & man & Mann & homme & $\cdots$\\
       $\vdots$ & \\
       \bottomrule
  \end{tabular}
  \caption{Schematic illustration of concept lists.}
  
%   \todo[inline]{SE: move to appendix if we need space}}
  \label{tab:swadesh}
\end{table}

\section{Family codes list}

\begin{table}[!htb]
    \centering
    \begin{tabular}{l|c}
    \toprule
    Family Code & Family Name \\
    \midrule
        AA & Afro-Asiatic\\
An & Austronesian \\
AuA & Austro-Asiatic \\
Bas & Basque \\
Dra & Dravidian \\
IE & Indo-European \\
Jap & Japonic \\
Kor & Koreanic\\
Krt & Kartvelian \\
NC & Niger-Congo \\
NDa & Nakh-Dagestan\\
ST & Sino-Tibetan \\
Tur & Turkic \\
Ura & Uralic\\
\bottomrule
    \end{tabular}
    \caption{Family codes list}
    \label{tab:famcodes}
\end{table}

\end{document}